\documentclass[final]{cvpr}

\usepackage{times}
\usepackage{epsfig}
\usepackage{graphicx}
\usepackage{amsmath}
\usepackage{amssymb}

\usepackage{subcaption}
\usepackage{booktabs}
\usepackage{multirow}
\usepackage{amsthm}
\usepackage[flushleft]{threeparttable}
\usepackage{colortbl}

\makeatletter
\@namedef{ver@everyshi.sty}{}
\makeatother
\usepackage{tkz-euclide}
\usepackage{pgfplots}
\pgfplotsset{compat=1.14}
\usetikzlibrary{positioning,calc,fadings}
\usetikzlibrary{backgrounds}

\usepackage{subcaption}
\usepackage{dsfont}
\usepackage{pifont}

\usepackage[pagebackref=true,breaklinks=true,colorlinks,bookmarks=false]{hyperref}

\DeclareMathOperator*{\argmax}{arg\,max}
\newtheorem*{definition*}{Definition}

\def\confYear{CVPR 2021}

\begin{document}

\title{Point Cloud Instance Segmentation using Probabilistic Embeddings} 

\author{Biao Zhang\\
KAUST\\
{\tt\small biao.zhang@kaust.edu.sa}
\and
Peter Wonka\\
KAUST\\
{\tt\small pwonka@gmail.com}
}

\maketitle
\begin{abstract}
     In this paper we propose a new framework for point cloud instance segmentation. Our framework has two steps: an embedding step and a clustering step. In the embedding step, our main contribution is to propose a probabilistic embedding space for point cloud embedding. Specifically, each point is represented as a tri-variate normal distribution.
     In the clustering step, we propose a novel loss function, 
     which benefits both the semantic segmentation and the clustering.
     Our experimental results show important improvements to the SOTA, i.e., 3.1\% increased average per-category mAP on the PartNet dataset.
\end{abstract}

\section{Introduction}

In this paper we tackle the problem of instance segmentation of point clouds. In instance segmentation we would like to assign two labels to each point in a point cloud. The first label is the class label (\eg, leg, back, seat, ... for a chair data set) and the second label is the instance ID (a unique number, \eg, to distinguish the different legs of a chair).
While instance segmentation had many recent successes in the image domain~\cite{he2017mask,liu2018path,fathi2017semantic,novotny2018semi,de2017semantic,neven2019instance}, we believe that the problem of instance segmentation for point clouds is not sufficiently explored.

We build our work on the idea of embedding-based instance segmentation, that is very popular in the image and volume domain~\cite{fathi2017semantic,novotny2018semi,de2017semantic,neven2019instance,lahoud20193d} and has also been successfully applied in the point clouds domain~\cite{wang2018sgpn,wang2019associatively}. In this approach typically two steps are employed. In the first step, each point (or pixel) is embedded in a feature space such that points belonging to the same instance should be close and points belonging to different instances should be further apart from each other. In the second step points are grouped using a clustering algorithm, such as mean-shift or greedy clustering. 

One important design choice in embedding-based methods is the dimensionality of the feature space. Some methods propose to use a high dimensional feature space~\cite{de2017semantic,kong2018recurrent}, while others use a low dimensional features space that has the same dimensionality as the input data~\cite{novotny2018semi,kendall2018multi,neven2019instance}, \eg, 2D for images, and 3D for point clouds. 
Methods with a low dimensional embedding space not only have lower computational complexity,
but they also lead to better interpretability, \eg,
embeddings are encoded as offset vectors towards instance centers.

Therefore, the main goal of our work is to extend the expressiveness of the embedding space in a way that leads to improved segmentation performance. Our proposed solution is to employ probabilistic embeddings, such that each point in the embedding space is encoded by a distribution. While assessing uncertainty is a popular tool in recent computer vision research~\cite{kendall2017uncertainties,Khan_2019_CVPR,Liu_2019_CVPR,Dorta_2018_CVPR} and we introduce this idea to the task of instance segmentation.
Incorporating uncertainty leads to an important improvement in segmentation performance. 
For example, on the PartNet~\cite{mo2019partnet} fine-grained instance segmentation dataset we can improve the SOTA by 3.1\% average per-category mAP.

In the remainder of the paper, we will give more details on the probabilistic embedding algorithm (Sec.~\ref{sec:method}), explain the embedding step (Sec.~\ref{sec:prob-embed}) and the clustering step (Sec.~\ref{sec:sem-cls}) in more detail.

\pgfmathdeclarefunction{gauss}{2}{%
  \pgfmathparse{1/(#2*sqrt(2*pi))*exp(-((x-#1)^2)/(2*#2^2))}%
}

\tikzfading
[
  name=fade out,
  inner color=transparent!0,
  outer color=transparent!100
]
\tikzset{rv/.style={red,path fading=fade out,}}

\tikzfading
[
  name=fade out,
  inner color=transparent!0,
  outer color=transparent!100
]
\tikzset{rv/.style={red,path fading=fade out,}}

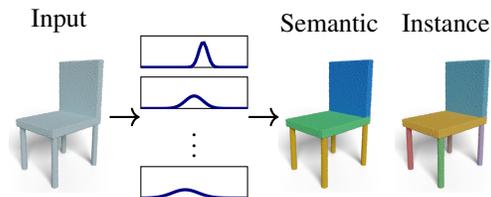
\begin{figure}[tb]
    \centering
    \begin{tikzpicture}[remember picture]
    \pgfplotsset{every axis/.style={
            enlargelimits=false,
            no markers,
            domain=-6:6,
            samples=50,
            xtick=\empty,
            ytick=\empty,
            ymax=1,
            ymin=0,
            height=2cm, 
            width=3cm,
        }}

        \node[inner sep=0pt, label={Input}] (input) {\includegraphics[width=.16\linewidth]{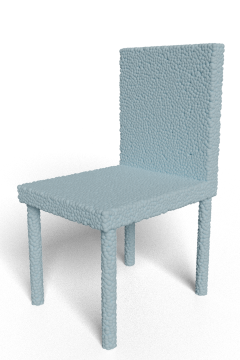}};
        
        \node[inner sep=0pt, right = 0.4cm of input] (embed) {
            \begin{tikzpicture}
                \node[] (A) {
                    \begin{tikzpicture}
                        \begin{axis}
                        \addplot [very thick,blue!50!black] {gauss(1,0.5)};
                        \end{axis}
                    \end{tikzpicture}};
                \node[inner sep=0pt, below = 0.1cm of A] (B) {
                    \begin{tikzpicture}
                        \begin{axis}
                        \addplot [very thick,blue!50!black] {gauss(0,1.0)};
                        \end{axis}
                    \end{tikzpicture}};
                \node[inner sep=0pt, below = 0.1cm of B] (C) {\vdots};
                \node[inner sep=0pt, below = 0.1cm of C] (D) {
                    \begin{tikzpicture}
                        \begin{axis}
                        \addplot [very thick,blue!50!black] {gauss(-1,1.5)};
                        \end{axis}
                    \end{tikzpicture}
                };
            \end{tikzpicture}};
            
                \node[right = 0.4cm of embed, inner sep=0pt, label={Semantic}] (sem) {\includegraphics[width=0.16\linewidth]{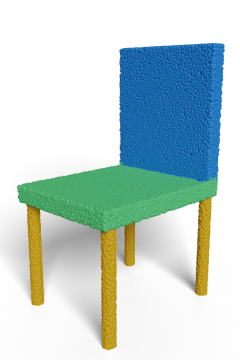}};
        
                \node[right = 0.2cm of sem, inner sep=0pt, label={Instance}] (ins) {\includegraphics[width=0.16\linewidth]{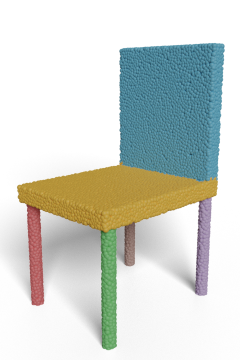}};
        
        \draw[thick,->] (input) -- (embed);
        \draw[thick,->] (embed) -- (sem);
    \end{tikzpicture}
    \vspace{-8pt}
    \caption{Our method takes a point cloud as input, encodes the points as random variables, and outputs semantic class labels and instance labels.}
    \label{fig:teaser}
\end{figure}
    
\begin{figure}
        \centering
    \begin{subfigure}{0.33\linewidth}
        \begin{tikzpicture}
            \begin{axis}[
                scale only axis,
                width=0.8\textwidth,
                height=0.8\textwidth,
                xmin=0,xmax=3,
                ymin=0,ymax=3,
                xticklabel=\empty,
                yticklabel=\empty,
                axis lines = middle
            ]
            \addplot[color = black, mark=*] coordinates {
            (1, 2)
            (2, 1)
            } node[above, pos=0]{$\mathbf{e}_i$} node[below, pos=1.0]{$\mathbf{e}_j$};
            \end{axis}
        \end{tikzpicture}
        \vspace{-10pt}
        \caption{Euclidean}
    \end{subfigure}%
    \begin{subfigure}{0.33\linewidth}
        \begin{tikzpicture}
            \begin{axis}[
                scale only axis,
                width=0.8\textwidth,
                height=0.8\textwidth,
                xmin=0,xmax=3,
                ymin=0,ymax=3,
                xticklabel=\empty,
                yticklabel=\empty,
                axis lines = middle
            ]
            \addplot[color = black, mark=*] coordinates {
            (0, 0)
            (1, 2)
            } node[above, pos=1]{$\mathbf{e}_i$};
            \addplot[color = black, mark=*] coordinates {
            (0, 0)
            (2, 1)
            } node[below, pos=1]{$\mathbf{e}_j$};
            \path 
                (1, 0.5)node[](A){} 
                (0.5, 1)node[](B){}
                (0, 0)node[](O){};
            \end{axis}
            \tkzMarkAngle[](A,O,B);
        \end{tikzpicture}
        \vspace{-10pt}
        \caption{Cosine}
    \end{subfigure}%
    \begin{subfigure}{0.33\linewidth}
        \begin{tikzpicture}
            \begin{axis}[
                scale only axis,
                width=0.8\textwidth,
                height=0.8\textwidth,
                xmin=0,xmax=3,
                ymin=0,ymax=3,
                xticklabel=\empty,
                yticklabel=\empty,
                axis lines = middle
            ]
                \addplot[color = black, mark=*] coordinates {
                    (1, 2)
                } node[above, pos=1]{$\mathbf{e}_i$};
                \fill[rv] (1,2) ellipse [x radius=1, y radius=0.5];
                \addplot[color = black, mark=*] coordinates {
                    (2, 1)
                } node[below, pos=1]{$\mathbf{e}_j$};
                \fill[rv,color = blue] (2,1) ellipse [x radius=0.7, y radius=1.0];
            \end{axis}
        \end{tikzpicture}
        \vspace{-10pt}
        \caption{Probabilistic}
    \end{subfigure}%
    \vspace{-10pt}
    \caption{\textbf{Examples of (dis)similarity measures.}}
    \label{fig:euc-cos}
\end{figure}
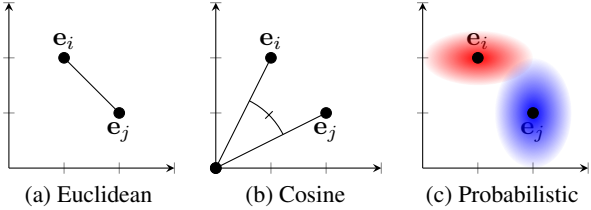

\paragraph{Contribution.} Our main contributions are as follows
\begin{enumerate}
    \item We propose to use probabilistic embeddings for instance segmentation and present a complete framework in the context of point cloud instance segmentation based on probabilistic embeddings.
    \item We develop a new loss function for the clustering step that is especially suited for high-granularity data sets.
    \item We show that the proposed probabilistic embeddings can be incorporated into existing embedding-based methods.
\end{enumerate}
\section{Related work}
\subsection{2D image instance segmentation}
The dominant approaches for image instance segmentation are proposal-based methods~\cite{he2017mask,liu2018path}, which are built upon object detection methods~\cite{girshick2015fast,ren2015faster}. Typically, they have higher quality, but a slower computation time compared to proposal free methods. The mainstream proposal free approaches are based on metric learning. 
The basic idea is to learn an embedding space, in which pixels belonging to the same object instance are close to each other and distant to pixels belonging to other object instances~\cite{fathi2017semantic,de2017semantic}. All above works are based on high-dimensional embedding, while more recent works~\cite{liang2017proposal,kendall2018multi,novotny2018semi,neven2019instance} show that 2D spatial embedding is sufficient to achieve the same or even higher performance.

\subsection{3D point cloud instance segmentation}
SGPN~\cite{wang2018sgpn} uses PointNet++~\cite{qi2017pointnet++} as backbone network and designs a double-hinge loss function to learn a pairwise similarity matrix of points. GSPN~\cite{yi2019gspn} produces object proposals with high objectness for point cloud instance segmentation. ASIS~\cite{wang2019associatively} is a module capable of making semantic segmentation and instance segmentation take advantage of each other. \cite{mo2019partnet} release a large scale point cloud dataset for part instance segmentation and benchmark their method and SGPN on this dataset. PointGroup~\cite{jiang2020pointgroup} and OccuSeg~\cite{han2020occuseg} achieved great success in scene datasets by voxelizing point clouds.

\subsection{Uncertainty in computer vision}
\cite{kendall2017uncertainties} present a unified framework combining model uncertainty with data uncertainty and can estimate uncertainty in classification and regression tasks. We introduce uncertainty estimation to the literature of instance embedding, by modeling points as random variables. Our method is related to recent works in deep generative networks~\cite{kingma2013auto,rezende2014stochastic}. They use a stochastic encoder to encode a data sample as a set of random variables, while focusing on solving the problem of backpropagation through random variables in deep neural networks. We deal with this problem by using a probabilistic product kernel~\cite{jebara2004probability}.

\section{Method}
\label{sec:method}

A training sample is a labeled 3D point cloud. It consists of point coordinates $\left\{\mathbf{x}_i\right\}^N_{i=1}$, class labels $\left\{y_i\right\}^N_{i=1}$ and instance IDs $\left\{z_i\right\}^N_{i=1}$.
We want to train a neural network to infer per point class labels and per point instance IDs at the same time.

\subsection{Probabilistic spatial embedding}
\label{sec:prob-embed}

A common approach in the literature of instance segmentation is to learn a function to embed pixels/points into a space where pair-wise similarity can be measured. Usually, this function is a deep neural network $f$ which transforms an unordered point set $\left\{\mathbf{x}_i\right\}^N_{i=1}$ to embeddings $\left\{\mathbf{e}_i\right\}^N_{i=1}$.

Instead of deterministic embeddings used in previous work, here we consider a probabilistic embedding, by modeling $\mathbf{e}_i$ as a random variable, $\mathbf{e}_i \sim p_i(\mathbf{e}),$

where $p_i$ is a probability density function. In Section~\ref{sec:ins-group} we will need to calculate the sum of random variables. In the ideal case, the distribution of a single random variable and the sum of multiple random variables has the same type of distribution that can be described with a few parameters. 
We choose to work with the tri-variate Gaussian distribution\footnote{Refer to \cite{liang2017proposal,kendall2018multi,novotny2018semi,neven2019instance} for a discussion why spatial embedding works.}
$p_i(\mathbf{e}) = \mathcal{N}(\mathbf{e};\boldsymbol{\mu}_i, \boldsymbol{\Sigma}_i)$

with mean vector $\boldsymbol{\mu}_i\in\mathbb{R}^3$ and covariance matrix $\boldsymbol{\Sigma}_i\in\mathbb{R}^{3\times 3}$.
For simplicity, let $\boldsymbol{\Sigma}_i$ be a diagonal matrix,
$\boldsymbol{\Sigma}_i = \mathop{\mathrm{diag}}(\sigma_i^{(1)2},\sigma_i^{(2)2},\sigma_i^{(3)2})$,
where $\sigma_i^{(d)2}$ is the square of $\sigma_i^{(d)}$ and $d=1,2,3$.

The network $f(\cdot)$ takes as input a (unordered) point set $\left\{\mathbf{x}_i\right\}^N_{i=1}$, and outputs $\left\{\boldsymbol{\mu}_i, \boldsymbol{\sigma}_i,\mathbf{p}_i\right\}^N_{i=1}$,
    $f(\left\{\mathbf{x}_i\right\}^N_{i=1}) = \left\{\boldsymbol{\mu}_i, \boldsymbol{\sigma}_i,\mathbf{p}_i\right\}^N_{i=1}$,
where $\boldsymbol{\sigma}_i = \left[\sigma_i^{(1)},\sigma_i^{(2)},\sigma_i^{(3)}\right]^{\intercal}\in\mathbb{R}^3$ and $\mathbf{p}_i$ is a probability vector which can be used to infer class label of $\mathbf{x}_i$ and will be explained in Sec.~\ref{sec:sem-cls}.

\subsection{Similarity measure}
\label{sec:sim}

In deterministic embeddings, the (dis)similarity between points is usually measured by Euclidean distance $\left\|\mathbf{e}_i-\mathbf{e}_j\right\|,$ or cosine similarity $\frac{\mathbf{e}_i^\intercal\mathbf{e}_j}{\left\|\mathbf{e}_i\right\|\left\|\mathbf{e}_j\right\|}$ (See Figure~\ref{fig:euc-cos}).
Since now we are using probabilistic embeddings, a similarity kernel for random variables needs to be selected. Here we describe the Bhattacharyya kernel~\cite{jebara2004probability}. 

\begin{definition*}
Let $\mathcal{P}$ be the set of distributions over $\Omega$. The Bhattacharyya kernel on $\mathcal{P}$ is the function $\mathcal{K}:\mathcal{P}\times\mathcal{P}\mapsto\mathbb{R}$ such that, for all $p, q \in \mathcal{P}$,
$\mathcal{K}(p, q) =  \int_{\Omega} \sqrt{p(\mathbf{x})} \sqrt{q(\mathbf{x})} \mathrm{d}\mathbf{x}$.   
\end{definition*}
We choose this kernel as our similarity measure for two reasons, 1) the Bhattacharyya kernel is symmetric, i.e. $\mathcal{K}(p, q)=\mathcal{K}(q, p)$; 2) the Bhattacharyya kernel has values between $0$ (no similarity) and $1$ (maximal similarity). And $\mathcal{K}(p, q)=1$ if and only if $p=q$.

Then the similarity $\kappa(\cdot, \cdot)$ between random variables can be represented by the Bhattacharyya kernel of their probability density functions\footnote{Refer to \cite{jebara2004probability} for a derivation.},
\begin{equation}
    \begin{aligned}
    \kappa(\mathbf{e}_i, \mathbf{e}_j)
    &= \int \sqrt{\mathcal{N}(\mathbf{e};\boldsymbol{\mu}_i, \boldsymbol{\Sigma}_i)} \sqrt{\mathcal{N}(\mathbf{e};\boldsymbol{\mu}_j, \boldsymbol{\Sigma}_j)} \mathrm{d}\mathbf{e} \\
    &= \beta_{i,j}\exp\left(-\left\|\boldsymbol{\mu}_i-\boldsymbol{\mu}_j\right\|^2_{\boldsymbol{\Sigma}_{i,j}^{-1}}\right),
    \end{aligned}
\end{equation}

where
\begin{align*}
\alpha_{i,j}^{(d)} &= 4(\sigma_i^{(d)2}+\sigma_j^{(d)2}), \\
\quad\beta_{i,j} &= \left(\prod^{3}_{d=1}\left(\sigma_i^{(d)}/\sigma_j^{(d)}+\sigma_j^{(d)}/\sigma_i^{(d)}\right)/2\right)^{-\frac{1}{2}},\\
\boldsymbol{\Sigma}_{i,j} &= \mathop{\mathrm{diag}}(\alpha_{i,j}^{(1)} ,\alpha_{i,j}^{(2)},\alpha_{i,j}^{(3)}),\\
 \left\|\boldsymbol{\mu}_i-\boldsymbol{\mu}_j\right\|_{\boldsymbol{\Sigma}^{-1}_{i,j}}^2 &= (\boldsymbol{\mu}_i-\boldsymbol{\mu}_j)^\intercal \boldsymbol{\Sigma}^{-1}_{i,j} (\boldsymbol{\mu}_i-\boldsymbol{\mu}_j) \\
 &= \sum^3_{d=1} (\mu_i^{(d)}-\mu_j^{(d)})/\alpha_{i,j}^{(d)}.
\end{align*}

\begin{itemize}
    \item If the uncertainties $\boldsymbol{\sigma}_i$ and $\boldsymbol{\sigma}_j$ have a large difference, $\beta_{i,j}$ will be small, so will be $\kappa(\mathbf{e}_i, \mathbf{e}_j)$. See Fig.~\ref{fig:sim}.
    \item If the centers $\boldsymbol{\mu}_i$ and $\boldsymbol{\mu}_j$ have a large difference, the exponential term will be small, so will be $\kappa(\mathbf{e}_i, \mathbf{e}_j)$. See Fig.~\ref{fig:sim}.
    \item The scale term $\beta_{i,j}=1$ if and only if the uncertainties $\boldsymbol{\sigma}_i$ and $\boldsymbol{\sigma}_j$ are element-wise equal. In this case, $\kappa(\mathbf{e}_i, \mathbf{e}_j)$ becomes an anisotropic Gaussian kernel, \begin{equation}
    \kappa_{RBF}(\boldsymbol{\mu}_i, \boldsymbol{\mu}_j) = \exp\left(-\left\|\boldsymbol{\mu}_i-\boldsymbol{\mu}_j\right\|^2_{\boldsymbol{\Sigma}_{i,j}^{-1}}\right).\label{eq:rbf}\end{equation}
    \item The exponential term equals $1$ if and only if the centers $\boldsymbol{\mu}_i$ and $\boldsymbol{\mu}_j$ are element-wise equal. In this case, $\kappa(\mathbf{e}_i, \mathbf{e}_j)$ becomes $\beta_{i,j}$, i.e., the similarity between uncertainties. This property allows two points that have the same embedding centers to have a low similarity, as long as $\beta_{i,j}$ is small.
\end{itemize}

\begin{figure}[tb]
    \centering
    \includegraphics[width=0.9\columnwidth]{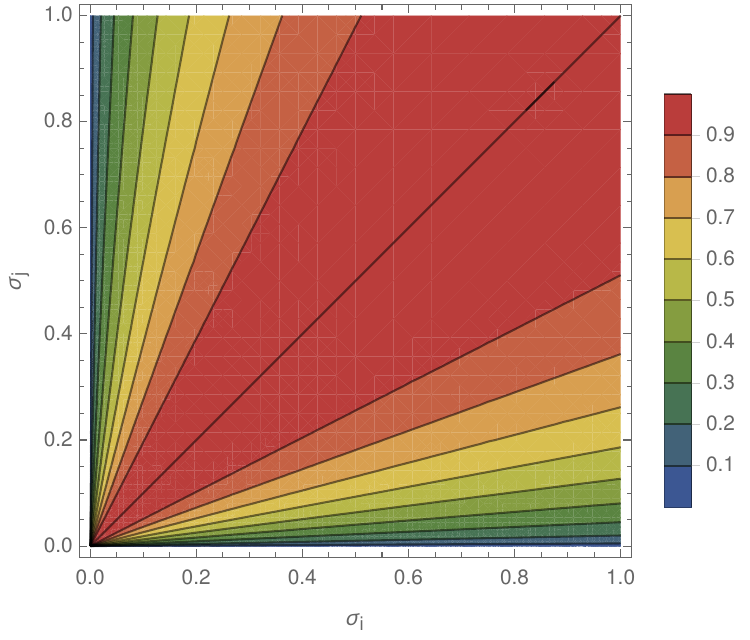}
    \vspace{-10pt}
    \caption{Contour plot of the 1-D uncertainty similarity $\beta_{i,j}=\left(\left(\sigma_i/\sigma_j+\sigma_j/\sigma_i\right)/2\right)^{-1/2}$. The highest value $1$ is achieved when $\sigma_i=\sigma_j$. The value goes to $0$ when one of the uncertainties is small and the other is large.}
    \label{fig:beta}
\end{figure}
\begin{figure}[tb]
    \centering
    \begin{subfigure}{0.4\linewidth}
        \includegraphics[width=\textwidth]{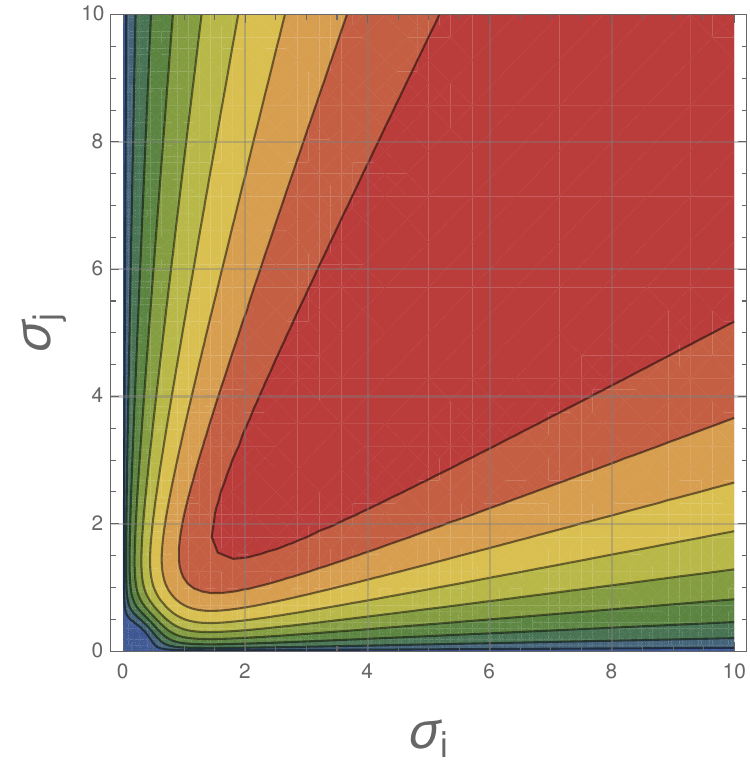}
    \end{subfigure}%
    \begin{subfigure}{0.4\linewidth}
        \includegraphics[width=\textwidth]{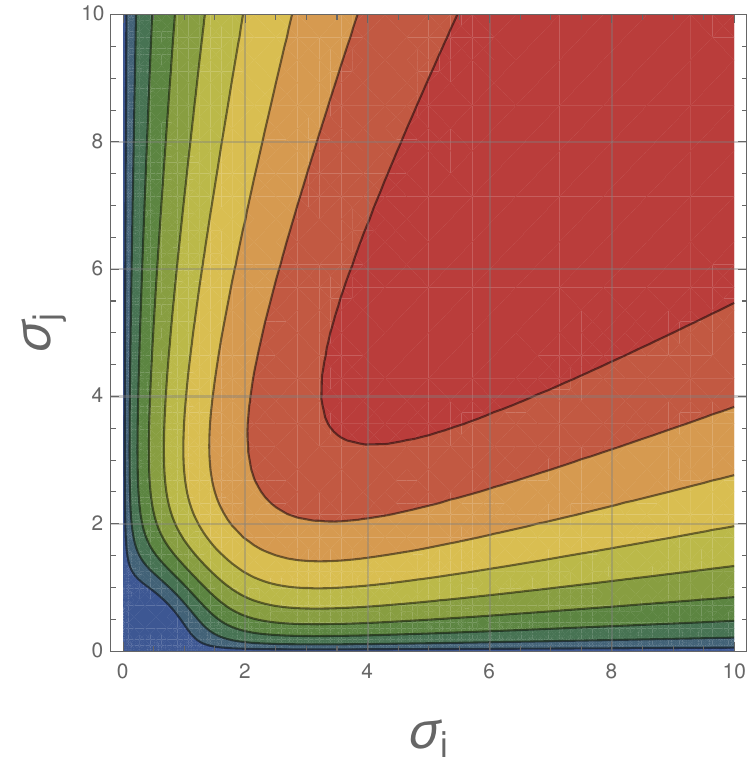}
    \end{subfigure}
    \begin{subfigure}{0.4\linewidth}
        \includegraphics[width=\textwidth]{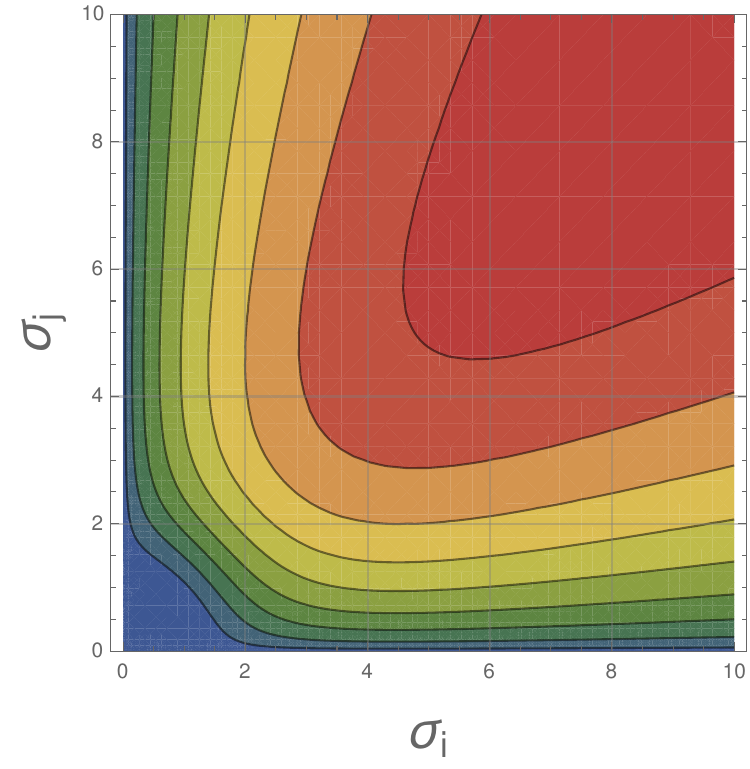}
    \end{subfigure}%
    \begin{subfigure}{0.4\linewidth}
        \includegraphics[width=\textwidth]{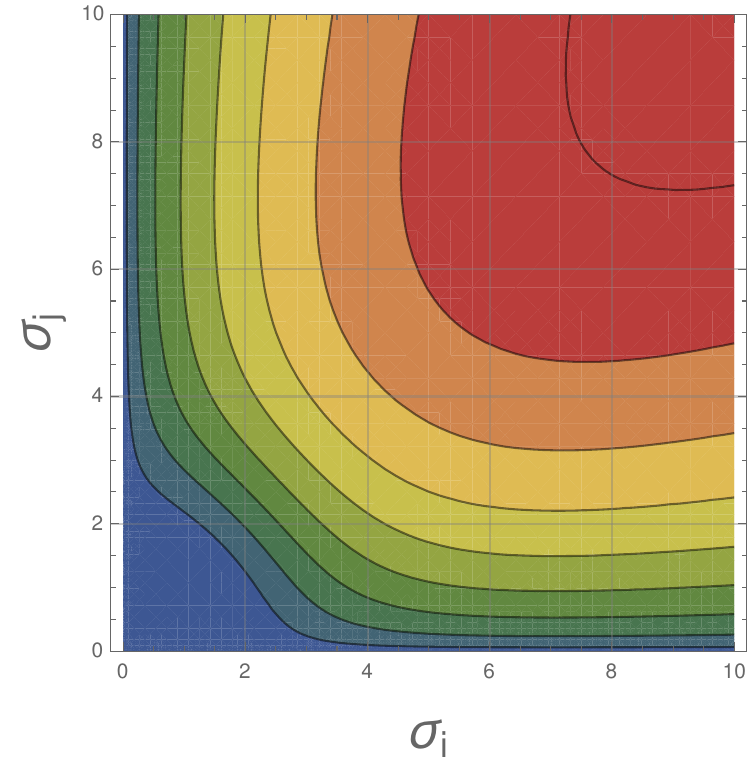}
    \end{subfigure}
    \vspace{-10pt}
    \caption{Contour plots of the similarity for different values of $\Delta \mu$ (from left to right, from top to below, $\Delta \mu=1,5,10,25$) in the case of 1-D probabilistic embedding. The similarity becomes $\beta_{i,j}\exp\left(-\frac{1}{2}\frac{\Delta \mu}{\alpha_{i,j}}\right)$, where $\Delta \mu = \left(\mu_i - \mu_j\right)^2$, $\alpha_{i,j}=4\left(\sigma_i^2+\sigma_j^2\right)$. The legend is the same as in Figure~\ref{fig:beta}.}
    \label{fig:sim}

\end{figure}

Compared to deterministic embedding, $\kappa(\mathbf{e}_i,\mathbf{e}_j)=\exp \left(-\left\|\mathbf{e}_i-\mathbf{e}_j\right\|^2\right)$
our similarity measure consists not merely of the similarity of spatial distances, but also the similarity of uncertainties.

In the following, we discuss multiple choices of embedding distributions that we will evaluate in Sec~\ref{sec:ablation}.

\paragraph{Homoscedasticity vs. Heteroscedasticity.}
The embeddings $\{\mathbf{e}_i\}^{N}_{i=1}$ are homoscedastic if they have the same variance $\boldsymbol{\Sigma}$. In this case, for a point cloud $\mathbf{X}$ we learn to predict a single $\boldsymbol{\Sigma}$ instead of point-dependent variances $\left\{\boldsymbol{\Sigma}_i\right\}^N_{i=1}$. And the similarity kernel becomes 
$\kappa(\mathbf{e}_i,\mathbf{e}_j) = \exp\left(-\left\|\boldsymbol{\mu}_i-\boldsymbol{\mu}_j\right\|^2_{\boldsymbol{\Sigma}_{i,j}^{-1}}\right)$,
which is also the form of the RBF kernel in Eq.~\eqref{eq:rbf}.

\paragraph{Isotropy vs. Anisotropy.}
The variance $\boldsymbol{\Sigma}_i$ is isotropic if its diagonal elements (variances of dimensions) are the same. Then we can write
$\boldsymbol{\Sigma}_i=\sigma_i^2\mathbf{I}$, where $\mathbf{I}$ is a $3\times 3$ identity matrix. The similarity can be written as 
\begin{equation}
    \kappa(\mathbf{e}_i,\mathbf{e}_j) = \beta_{i, j} \exp\left(-\frac{\left\|\boldsymbol{\mu}_i-\boldsymbol{\mu}_j\right\|^2}{\alpha_{i,j}}\right),
\end{equation}
where 
$
    \beta_{i,j} = \left(\left(\sigma_i/\sigma_j+\sigma_j/\sigma_i\right)/2\right)^{-\frac{3}{2}}$ and $\alpha_{i,j}=4(\sigma_i^2+\sigma_j^2)$.
\subsection{Instance grouping}
\label{sec:ins-group}

Let $\left\{i:z_i=k\right\}$ be the index set of points having instance ID $k$. We take an average of these embeddings to get the embedding $\mathbf{c}_k$ of instance $k$,
    $\mathbf{c}_k = \frac{1}{\left|\{i:z_i=k\}\right|}\sum_{\{i:z_i=k\}} \mathbf{e}_i$.
Since the sum of Gaussian random variables is still a Gaussian random variable, we can derive the following: 
\begin{subequations}
\begin{align}
    p(\mathbf{c}_k) &= \mathcal{N}(\mathbf{c}_k;\hat{\boldsymbol{\mu}}_k, \hat{\boldsymbol{\Sigma}}_k), \\
    \hat{\boldsymbol{\mu}}_k &= \frac{1}{\left|\{i:z_i=k\}\right|}\sum_{\{i:z_i=k\}} \boldsymbol{\mu}_i, \label{eqn:ins-m}\\
    \hat{\boldsymbol{\Sigma}}_k &= \frac{1}{\left|\{i:z_i=k\}\right|}\sum_{\{i:z_i=k\}} \boldsymbol{\Sigma}_i.\label{eqn:ins-s}
\end{align}
\end{subequations}
Now we can measure the similarity between a point and an instance by using $\kappa(\mathbf{e}_i, \mathbf{c}_k)$.

If $z_i=k$, we want $\kappa(\mathbf{e}_i, \mathbf{c}_k)$ to be close to 1, otherwise 0. We can optimize a binary cross entropy loss function,
\begin{equation}
    \mathcal{L}_{InsCE} = \frac{1}{NK}\sum^{K-1}_{k=0}\sum^{N}_{i=1}
    \left\{\begin{array}{lr}
        -\ln \kappa(\mathbf{e}_i,\mathbf{c}_k), & \text{if } z_i=k,\\
        -\ln (1-\kappa(\mathbf{e}_i,\mathbf{c}_k)), & \text{otherwise.}
    \end{array}\right.
\end{equation}
However, in practice, this suffers from a serious foreground-background imbalance problem. To remedy this drawback we propose to use the combined log-Dice loss function~\cite{wong20183d} instead:
\begin{equation}
    \mathcal{L}_{Ins} = \mathcal{L}_{InsCE} -\ln \frac{2\sum_{k=0}^{K-1}\sum^N_{i=1}\kappa(\mathbf{e}_i, \mathbf{c}_k)\mathds{1}_{z_i=k}}{\sum_{k=0}^{K-1}\sum^N_{i=1}\left(\kappa(\mathbf{e}_i, \mathbf{c}_k) + \mathds{1}_{z_i=k}\right)},
\end{equation}
where $\mathds{1}_{z_i=k}$ is an indicator function which equals $1$ when $z_i=k$, $0$ otherwise.

\paragraph{Entropy Regularization.}
As we can see in Figure~\ref{fig:sim}, when $\sigma_i^{(l)}$ and $\sigma_j^{(l)}$ goes to infinity while keeping $\sigma_i^{(l)}=\sigma_j^{(l)}$, $\beta_{i,j}=1$ and the similarity equals to 1 no matter what the value $\boldsymbol{\mu}_i-\boldsymbol{\mu}_j$ is. Formally speaking,
\begin{equation}
    \lim_{\sigma_i^{(l)}\to\infty,\sigma_j^{(l)}\to\infty, \sigma_i^{(l)}=\sigma_j^{(l)},l=1,2,3}\kappa(\mathbf{e}_i, \mathbf{e}_j) = 1.
\end{equation}
Consequently, the similarity degenerates to constant $1$ for every pair of embeddings. To address this issue, we propose an entropy regularizer,
\begin{equation}
    \mathcal{L}_{Reg} = \sum^N_{i=1}\mathbb{H}(\mathbf{e}_i)=\sum^N_{i=1}\left[\frac{3}{2}\ln(2\pi e) + \frac{1}{2}\ln \left(\sigma_i^{(1)}\sigma_i^{(2)}\sigma_i^{(3)}\right)^2\right],
\end{equation}
where $\mathbb{H}(\mathbf{e}_i)$ is the entropy of multivariate Gaussian variable $\mathbf{e}_i$. This regularizer is not only able to prevent the similarity degeneration by minimizing the variances along all dimensions, but can also penalize large uncertainties, thus increasing the confidence of the network output as in~\cite{grandvalet2005semi} and \cite {wang2019dynamic}.

\subsection{Semantic classification}
\label{sec:sem-cls}

\begin{figure}[tb]
    \centering
    \begin{tikzpicture}[remember picture]
        \node[inner sep=0pt] (i1) {\includegraphics[width=.15\linewidth]{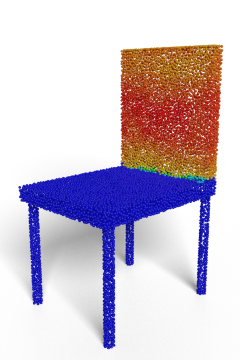}};
        \node[right = 0.1cm of i1, inner sep=0pt] (i2) {\includegraphics[width=.15\linewidth]{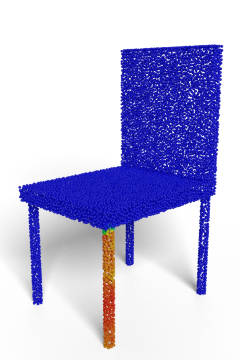}};
        \node[right = 0.1cm of i2, inner sep=0pt] (i3) {\includegraphics[width=.15\linewidth]{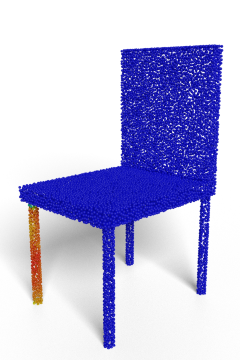}};
        \node[right = 0.1cm of i3, inner sep=0pt] (i4) {\includegraphics[width=.15\linewidth]{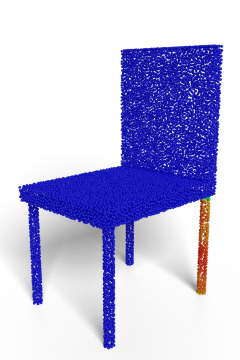}};
        \node[right = 0.1cm of i4, inner sep=0pt] (i5) {\includegraphics[width=.15\linewidth]{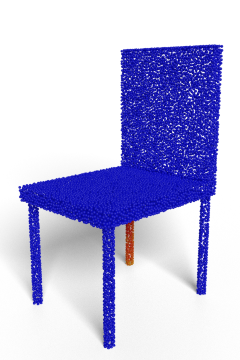}};
        \node[right = 0.1cm of i5, inner sep=0pt] (i6) {\includegraphics[width=.15\linewidth]{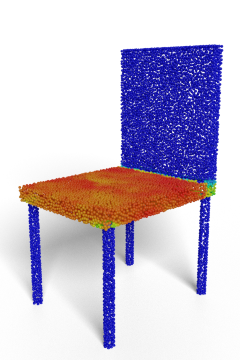}};
        \node[below = 1.2cm of $(i1)!0.5!(i2)$, inner sep=0pt] (c1) {\includegraphics[width=.33\linewidth]{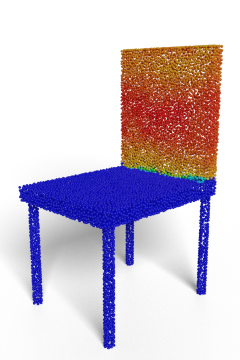}};
        \node[below = 1.2cm of $(i3)!0.5!(i4)$, inner sep=0pt] (c2) {\includegraphics[width=.33\linewidth]{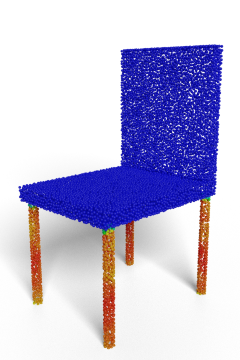}};
        \node[below = 1.2cm of $(i5)!0.5!(i6)$, inner sep=0pt] (c3) {\includegraphics[width=.33\linewidth]{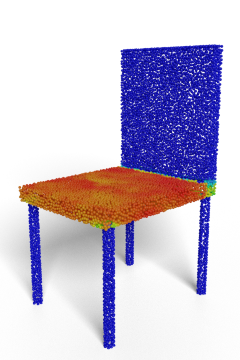}};
        
        \draw[thick,->] (i1) -- (c1);
        \draw[thick,->] (i2) -- (c2);
        \draw[thick,->] (i3) -- (c2);
        \draw[thick,->] (i4) -- (c2);
        \draw[thick,->] (i5) -- (c2);
        \draw[thick,->] (i6) -- (c3);
        
    \end{tikzpicture}
    \vspace{-10pt}
    \caption{\textbf{Top row}: similarity maps for each part instance $\tilde{\mathbf{Q}}[:, k]$. \textbf{Bottom row}: probability map $\mathbf{Q}[:, l]$. The arrows show that information of instances of the same class is aggregated in Eq.~\ref{eq:combine}. Red means high similarity, while blue means low.}
    \label{fig:score-map}
\end{figure}
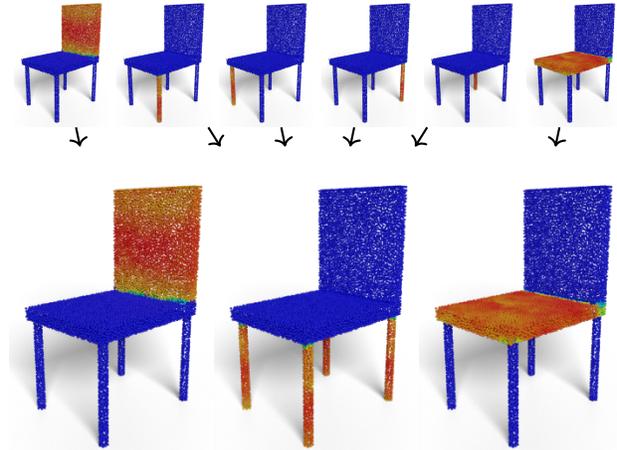

Neven \etal \cite{neven2019instance} introduces a way to use score maps to find cluster centers. Our main novelty is the new loss function, so our description focuses on this part. We still describe the greedy clustering steps from \cite{neven2019instance} for completeness. In Section~\ref{sec:ablation}, we compare our new \emph{center-aware loss} to the previously used MSE loss in~\cite{neven2019instance}.

After defining the similarity measure, we can easily find out all points similar to an instance center. However, during the inference phase, we don't have the information of ground-truth instance IDs, thus, it is impossible to use Eqs.~\ref{eqn:ins-m} and~\ref{eqn:ins-s} to get instance centers. Therefore, along with distribution parameters $\left\{\boldsymbol{\mu}_i\right\}^N_{i=1}$ and $\left\{\boldsymbol{\sigma}_i\right\}^N_{i=1}$, we also predict a score map $\left\{\mathbf{p}_i\right\}^N_{i=1}$, where $\mathbf{p}_i\in\mathbb{R}^{L}$ and its $l$-th entry $\mathbf{p}_i^{(l)}$ indicates the probability of $\mathbf{x}_i$ being an instance center with class label $l$. Thus we want $\mathbf{P}\in\mathbb{R}^{N\times L}$, the matrix form of $\left\{\mathbf{p}_i\right\}^N_{i=1}$, to satisfy two conditions:

\begin{enumerate}
    \item 
    $\mathbf{P}[i, :]$ is a probability vector and can be used to infer class label $y_i$ of point $\mathbf{x}_i$, i.e., $y_i = \arg\max_{l=1}^{L}\mathbf{P}[i, l]$.
    \item For foreground class labels $l\in\{1, 2, \dots, L\}$, $\mathbf{P}[:, l]$ is a score map of being an instance center with class label $l$.
\end{enumerate}

The first condition is easy to satisfy with the cross entropy loss. Assuming $\mathbf{P}[i,:]$ is the output of a softmax function, we can minimize,
\begin{equation}
    \label{eqn:clsce}
    \mathcal{L}_{ClsCE} = \frac{1}{N}\sum^N_{i=1}\left(\frac{1}{L}\sum^L_{l=1}-\mathds{1}_{y_i=l}\log \mathbf{P}[i,l]\right),
\end{equation}
where $\mathds{1}_{y_i=l}$ is an indicator function which equals $1$ when $y_i=l$, $0$ otherwise.

For the second condition, we take into account $\kappa(\mathbf{e}_i, \mathbf{c}_k)$, which is the similarity between $\mathbf{x}_i$ and an instance $k$. Consider $\tilde{\mathbf{Q}}\in\mathbb{R}^{N\times K}$, 
\begin{equation}
    \tilde{\mathbf{Q}}[i,k]=\kappa(\mathbf{e}_i, \mathbf{c}_k)\mathds{1}_{z_i=k}=\begin{cases}\kappa(\mathbf{e}_i, \mathbf{c}_k)&z_i=k,\\0&\text{otherwise},\end{cases}
\end{equation}
where each entry $\tilde{\mathbf{Q}}[i,k]$ can be interpreted as the probability of $\mathbf{x}_i$ being the center of instance $k$. Upon this we calculate $\mathbf{Q}\in\mathbb{R}^{N\times L}$, where $\mathbf{Q}[i,l]$ gives the probability of $\mathbf{x}_i$ being an instance center with class label $l$,
\begin{equation}
    \label{eq:combine}
    \mathbf{Q}[i,l]=\max_{\left\{k: y(k)=l\right\}} \tilde{\mathbf{Q}}[i,k],
\end{equation}
where $y(k)$ is the class label of instance $k$, due to the fact that $\left\{\mathbf{x}_i:z_i=k\right\}$ must have the same class label. (See an illustration in Figure~\ref{fig:score-map}.) After that, we want both $\mathbf{P}[:,l]$ and $\mathbf{Q}[:,l]$ to achieve local maxima at the same points for all $l\in \{1,2,\dots,L\}$. When we are doing inference, these local maxima are chosen as instance centers. Therefore, the first condition can be weakened, and only points which are close to instance centers should be classified correctly.

We design a new loss function to satisfy the two conditions at the same time,
\begin{equation}
    \mathcal{L}_{Score} = \frac{1}{NL}\sum_{i=1}^N\sum_{l=1}^L -\mathbf{Q}[i, l] \log \mathbf{P}[i, l].
    \label{eq:score}
\end{equation}
Here $\mathbf{Q}$ is fixed as a target when training.
We can view $\mathcal{L}$ in two ways,
\begin{enumerate}
    \item First, we switch the order of summation in Eq.~\eqref{eq:score},
    \begin{equation}
        \mathcal{L}_{Score} = \frac{1}{L}\sum_{l=1}^L\left(\frac{1}{N}\sum_{i=1}^N-\mathbf{Q}[i, l] \log \mathbf{P}[i, l]\right).
    \end{equation}
    The value of this quantity $-\mathbf{Q}[i, l]\log\mathbf{P}[i, l]$ is high when weight term $\mathbf{Q}[i, l]$ is high, and if we minimize it, we are forcing $-\log\mathbf{P}[i, l]$ to be small. Consequently, $\mathbf{P}[i, l]$ would be large. 
    This guarantees local maxima of $\mathbf{Q}[:, l]$ are also local maxima of $\mathbf{P}[:, l]$. And minimizing this loss term is equivalent to minimize the KL-divergence between (unnormalized probability) $\mathbf{Q}[:, l]$ and (unnormalized probability) $\mathbf{P}[:, l]$,
    \begin{equation}
        \mathbb{KL}(\mathbf{Q}[:, l]|\mathbf{P}[:, l]) = \frac{1}{N}\sum_{i=1}^N\mathbf{Q}[i, l] \log \frac{\mathbf{Q}[i, l]}{\mathbf{P}[i, l]}.
    \end{equation}
    \item Second, we look at the inner summation of Eq.~\eqref{eq:score},
    \begin{equation}
        \mathcal{L}_{Score} = \frac{1}{N}\sum_{i=1}^N\left(\frac{1}{L}\sum_{l=1}^L-\mathbf{Q}[i, l] \log \mathbf{P}[i, l]\right).
    \end{equation}
    The inner summation inside the round bracket is the cross entropy between $\mathbf{Q}[i, :]$ and $\mathbf{P}[i, :]$. And it is equivalent to replacing the one-hot vector in Equation \ref{eqn:clsce} with $\mathbf{Q}[i, :]$. Also, it is the form of label smoothing, a commonly used training trick in image classification~\cite{szegedy2016rethinking,he2019bag}. 
    The closer $\mathbf{Q}[i, :]$ is to a one-hot vector, the more confidence we give to the classification loss of point $\mathbf{x}_i$. By definition of $\mathbf{Q}[i, l]$, it can be easily seen that the resulting classifier only classifies near-centers points correctly. Thus we call our new loss function the \emph{center-aware loss}.
\end{enumerate}
\begin{table*}[htbp]
\small\addtolength{\tabcolsep}{-4.7pt}

\resizebox{\textwidth}{!}{%

\begin{tabular}{@{}l|c|l|cccccccccccccccccccccccc@{}}
\toprule
 &     & Avg  & \rotatebox[origin=lB]{90}{Bag}  & \rotatebox[origin=lB]{90}{Bed}  & \rotatebox[origin=lB]{90}{Bottle} & \rotatebox[origin=lB]{90}{Bowl} & \rotatebox[origin=lB]{90}{Chair} & \rotatebox[origin=lB]{90}{Clock} & \rotatebox[origin=lB]{90}{Dish} & \rotatebox[origin=lB]{90}{Disp} & \rotatebox[origin=lB]{90}{Door} & \rotatebox[origin=lB]{90}{Ear}  & \rotatebox[origin=lB]{90}{Faucet} & \rotatebox[origin=lB]{90}{Hat}  & \rotatebox[origin=lB]{90}{Key}  & \rotatebox[origin=lB]{90}{Knife} & \rotatebox[origin=lB]{90}{Lamp} & \rotatebox[origin=lB]{90}{Laptop}   & \rotatebox[origin=lB]{90}{Micro} & \rotatebox[origin=lB]{90}{Mug}  & \rotatebox[origin=lB]{90}{Fridge} & \rotatebox[origin=lB]{90}{Scis} & \rotatebox[origin=lB]{90}{Stora} & \rotatebox[origin=lB]{90}{Table} & \rotatebox[origin=lB]{90}{Trash} & \rotatebox[origin=lB]{90}{Vase} \\ \midrule\midrule
\multirow{4}{*}{\rotatebox[origin=c]{90}{SGPN}} & 1   & 55.7          & 38.8          & 29.8          & 61.9          & 56.9          & 72.4          & 20.3          & 72.2          & 89.3          & 49.0          & 57.8          & 63.2          & 68.7          & 20.0          & 63.2          & 32.7          & \textbf{100.0} & 50.6          & 82.2          & 50.6          & 71.7          & 32.9          & 49.2          & 56.8          & 46.6          \\
                  & 2   & 29.7          & -             & 15.4          & -             & -             & 25.4          & -             & 58.1          & -             & 25.4          & -             & -             & -             & -             & -             & 21.7          & \textbf{-}     & 49.4          & -             & 22.1          & -             & 30.5          & 18.9          & -             & -             \\
                  & 3   & 29.5          & -             & 11.8          & 45.1          & -             & 19.4          & 18.2          & 38.3          & 78.8          & 15.4          & 35.9          & 37.8          & -             & -             & 38.3          & 14.4          & \textbf{-}     & 32.7          & -             & 18.2          & -             & 21.5          & 14.6          & 24.9          & 36.5          \\ \cmidrule{2-27}
                  & Avg & 46.8          & 38.8          & 19.0          & 53.5          & 56.9          & 39.1          & 19.3          & 56.2          & 84.1          & 29.9          & 46.9          & 50.5          & 68.7          & 20.0          & 50.8          & 22.9          & \textbf{100.0} & 44.2          & 82.2          & 30.3          & 71.7          & 28.3          & 27.6          & 40.9          & 41.6          \\ \midrule
\multirow{4}{*}{\rotatebox[origin=c]{90}{PartNet}} & 1   & 62.6          & \textbf{64.7} & 48.4          & \textbf{63.6} & 59.7          & 74.4          & \textbf{42.8} & 76.3          & 93.3          & 52.9          & 57.7          & 69.6          & 70.9          & 43.9          & 58.4          & 37.2          & \textbf{100.0} & 50.0          & 86.0          & 50.0          & 80.9          & 45.2          & \textbf{54.2} & \textbf{71.7} & \textbf{49.8} \\
                  & 2   & 37.4          & -             & 23.0          & -             & -             & 35.5          & -             & 62.8          & -             & \textbf{39.7} & -             & -             & -             & -             & -             & 26.9          & \textbf{-}     & 47.8          & -             & 35.2          & -             & 35.0          & 31.0          & -             & -             \\
                  & 3   & 36.6          & -             & 15.0          & 48.6          & -             & 29.0          & \textbf{32.3} & \textbf{53.3} & 80.1          & 17.2          & 39.4          & 44.7          & -             & -             & \textbf{45.8} & 18.7          & \textbf{-}     & 34.8          & -             & 26.5          & -             & 27.5          & 23.9          & 33.7          & \textbf{52.0} \\ \cmidrule{2-27}
                  & Avg & 54.4          & \textbf{64.7} & 28.8          & 56.1          & 59.7          & 46.3          & \textbf{37.6} & \textbf{64.1} & 86.7          & 36.6          & 48.6          & 57.2          & 70.9          & 43.9          & 52.1          & 27.6          & \textbf{100.0} & 44.2          & 86.0          & 37.2          & 80.9          & 35.9          & \textbf{36.4} & 52.7          & \textbf{50.9} \\ \midrule\midrule
\multirow{4}{*}{\rotatebox[origin=c]{90}{\textbf{Ours}}} & 1   & \textbf{65.1} & 64.6          & \textbf{51.4} & 63.1          & \textbf{72.0} & \textbf{77.1} & 41.1          & \textbf{76.9} & \textbf{95.3} & \textbf{61.2} & \textbf{66.5} & \textbf{73.1} & \textbf{71.8} & \textbf{48.6} & \textbf{76.5} & \textbf{37.1} & \textbf{100.0} & \textbf{50.5} & \textbf{90.9} & \textbf{50.5} & \textbf{88.6} & \textbf{47.3} & 40.3          & 69.0          & 48.7          \\
                  & 2   & \textbf{40.4} & -             & \textbf{31.0} & -             & -             & \textbf{38.6} & -             & \textbf{64.2} & -             & 36.9          & -             & -             & -             & -             & -             & \textbf{31.0} & \textbf{-}     & \textbf{51.2} & -             & \textbf{37.3} & -             & \textbf{42.0} & \textbf{31.5} & -             & -             \\
                  & 3   & \textbf{39.8} & -             & \textbf{26.2} & \textbf{50.7} & -             & \textbf{34.7} & 30.2          & 50.0          & \textbf{82.0} & \textbf{25.7} & \textbf{43.2} & \textbf{55.6} & -             & -             & 44.4          & \textbf{20.3} & \textbf{-}     & \textbf{37.0} & -             & \textbf{31.1} & -             & \textbf{34.2} & \textbf{25.5} & \textbf{37.7} & 47.6          \\ \cmidrule{2-27}
                  & Avg & \textbf{57.5} & 64.6          & \textbf{36.2} & \textbf{56.9} & \textbf{72.0} & \textbf{50.1} & 35.6          & 63.7          & \textbf{88.7} & \textbf{41.3} & \textbf{54.9} & \textbf{64.4} & \textbf{71.8} & \textbf{48.6} & \textbf{60.5} & \textbf{29.5} & \textbf{100.0} & \textbf{46.2} & \textbf{90.9} & \textbf{39.6} & \textbf{88.6} & \textbf{41.2} & 32.4          & \textbf{53.4} & 48.1          \\ \bottomrule
                  
\end{tabular}
}
\vspace{-10pt}
\caption{\textbf{Instance segmentation results on PartNet (part-category mAP\%, IoU threshold 0.5, fine(3), middle(2), and coarse(1)-grained)}. 
}
\label{tbl:result}
\end{table*}

The inference process is done with a greedy approach \cite{neven2019instance}.
From foreground score maps $\left\{\mathbf{P}[:,1], \mathbf{P}[:,2], \dots, \mathbf{P}[:,L]\right\}$, we sample a point $\mathbf{x}_{i_0}$ with highest score $\mathbf{P}[i_0, l_0]$, where $i_0$ is the point index and $l_0$ is its class label,
\begin{equation}
    (i_0, l_0)=\argmax_{i\in\{1, 2, \dots, N\},\ l\in\{1, 2, \dots, L\}} \mathbf{P}[i, l].
\end{equation}
The point $\mathbf{x}_{i_0}$ is an anchor and we want to find all similar points. Specifically we find all points $\mathbf{x}_{i}$ with $\kappa(\mathbf{e}_i, \mathbf{e}_{i_0})\geq \tau.$ As a result, the instance ID of $\mathbf{x}_{i}$ is $0$. After that, all points satisfying the inequality are all masked out. Similarly, we sample $\mathbf{x}_{i_1}$ and mask out points with instance ID $1$, sample $\mathbf{x}_{i_2}$ and mask out points with instance ID $2$, and so on. We stop this loop if there is no point left. We use the validation set to fit hyperparameter $\tau$, which is $0.35$ in our experiments. 

\subsection{Implementation}
For a fair comparison to our main competitor PartNet~\cite{mo2019partnet} we keep as much of their structure as possible (Note that PartNet is the name of a dataset as well as an instance segmentation method). 
We also use PointNet++~\cite{qi2017pointnet++} as the feature extraction backbone, with the same parameters as~\cite{mo2019partnet}.
We use 3 output heads for centers, uncertainties, and scores as in $f(\left\{\mathbf{x}_i\right\}^N_{i=1}) = \left\{\boldsymbol{\mu}_i, \boldsymbol{\sigma}_i,\mathbf{p}_i\right\}^N_{i=1}$. We list the activation functions for output heads in Table~\ref{tbl:heads}. 

\begin{table}[tb]
        \begin{center}
        
        \begin{tabular}{@{}lcl@{}}
        \toprule
        & Output & Activation \\ \midrule
        Centers & $\mathbf{o}_i\in\mathbb{R}^3$ & $\boldsymbol{\mu}_i = \mathbf{x}_i+\tanh \mathbf{o}_i$ \\
        Uncertainties & $\tilde{\boldsymbol{\sigma}}_i\in\mathbb{R}^3$ & $\boldsymbol{\sigma}_i = \exp{\tilde{\boldsymbol{\sigma}}_i}$ \\
        Scores & $\tilde{\mathbf{p}}_i\in\mathbb{R}^3$ & $\mathbf{p}_i = \mathop{\mathrm{softmax}}(\tilde{\mathbf{p}}_i)$ \\ \bottomrule
        \end{tabular}
        \vspace{-8pt}
        \caption{\textbf{Activations for different output branches}}
        \label{tbl:heads}
        \end{center}
\end{table}

The final objective function is
\begin{equation}
    \mathcal{L} = \mathcal{L}_{Ins} + \mathcal{L}_{Score} + 0.001 \cdot \mathcal{L}_{Reg}
\end{equation}

We use random jittering, translation (between -0.01 and 0.01) and rotation (between $-15^\circ$ and $15^\circ$ for each axis) as data augmentation, and use the Adam~\cite{kingma2014adam} optimizer. We use a batch-size of $16$ and an initial learning rate of $0.001$ for $500$ epochs with a decay factor of $0.5$ at epoch 50 and epoch 150.
\begin{figure*}
    \centering
    \input{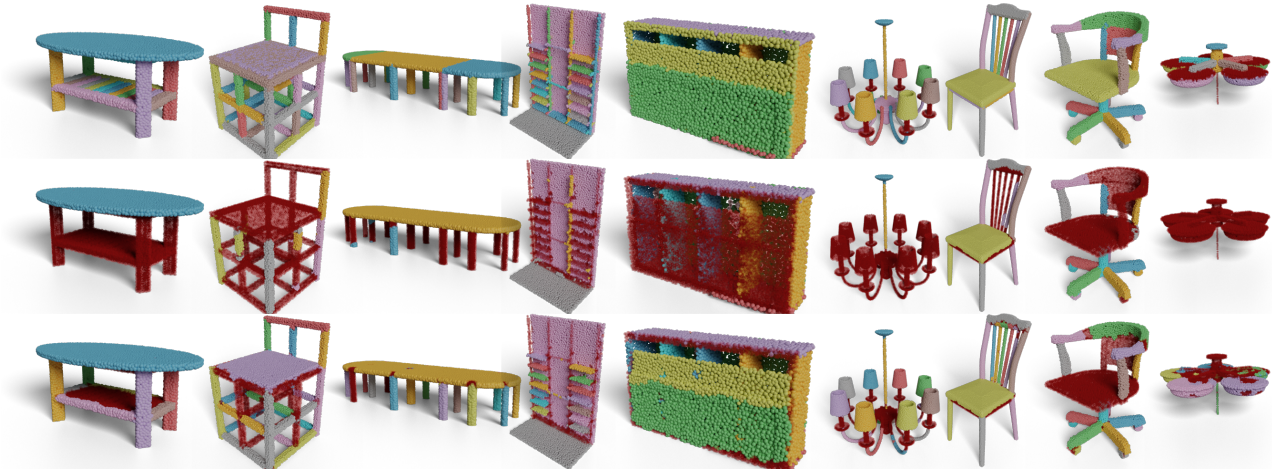}
    \vspace{-20pt}
    \caption{\textbf{Top row}: ground-truth. \textbf{Middle row}: PartNet. \textbf{Bottom row}: Probabilistic Embedding. We show true positives (IoU threshold 0.5) with the same color as ground truth. False detections are shown in \em{transparent red}.}
    \label{fig:results}
\end{figure*}
\section{Results}

PartNet~\cite{mo2019partnet} provides coarse-, middle- and fine-grained part instance-level annotations for 3D point clouds from ShapeNet~\cite{chang2015shapenet}. It contains 24 object categories, but the number of training samples varies greatly from 92 to 5707 for different categories.
In contrast to indoor scene point cloud datasets (\eg, ScanNet by \cite{dai2017scannet}), instances (object parts) of PartNet require more context to be classified and are connected. Many visually alike parts have different semantic labels, \eg, ping-pong table's legs and pool table's legs in the category of table. Also, instance masks should have no overlaps. All these make it a very challenging dataset for instance segmentation.

\subsection{Quantitative and qualitative results}

\paragraph{PartNet.} We report per-category mean Average Precision (mAP) scores for the PartNet dataset in Table~\ref{tbl:result}. The IoU threshold is $0.5$. 
We compare our probabilistic embedding algorithm to 
PartNet~\cite{mo2019partnet} and SGPN~\cite{wang2018sgpn}. The results are averaged over three levels of granularity (fine(3), middle(2), and coarse(1)). 

On the complete dataset, our method outperforms the best competitor PartNet by 3.1\% average per-category mAP. We can observe that our method has a slightly bigger advantage in fine-grained instance segmentation compared to coarse-grained instance segmentation (3.2\% vs. 2.5\%). We can also observe consistent improvements in categories with little as well as many training samples. While we beat SOTA in all categories with many training samples (Chair, Table, StorageFurniture, and Lamp), PartNet has better results in some of the categories with fewer training samples. 

We also show visualization examples in Figure~\ref{fig:results}. Compared to PartNet~\cite{mo2019partnet}, our method shows great improvement especially when there are many instances in a point cloud.

\paragraph{ScanNet.}
As baseline method we chose a network based on performance and availability of code. Since the best methods, such as OccuSeg, do not release code for ScanNet, we decided to build our own baseline using MinkowskiNet~\cite{choy20194d} as feature backbone. MinkowskiNet is a sparse tensor network that achieved great results on indoor semantic scene segmentation. In order to adapt the network to instance segmentation, we re-implemented the learnable margin method proposed by \cite{neven2019instance}. The learnable margin method does well on common image instance segmentation datasets and is well-balanced both in speed and accuracy.
This combination of two recent papers gives a strong baseline, but not state-of-the-art results in the metrics. We compare to this baseline, also using MinkowskiNet as feature backbone to make the results directly comparable.

We report the average precision (AP) in Table~\ref{tbl:scannet} and compare our method with other leading results on ScanNet. Although we do not have the overall state-of-the-art results, the improvement over the baseline \cite{neven2019instance} verifies the impact of probabilistic embedding and demonstrates that our method can be integrated with any embedding-based method and any backbone network. We can improve the validation mAP by $4.9\%$. We would also like to note that the main point of the paper is to showcase the benefit of the probabilisitc embedding method. We did not have the resources to fine-tune our method for the ScanNet dataset, but nevertheless our results are comparable with the state of the art and in some categories beating state of the art already. We therefore argue that this result underlines the significance of the proposed embedding method as it is likely that future state of the art methods will also be able to benefit from it.

\begin{table}[tb]
\centering
\small\addtolength{\tabcolsep}{-3.2pt}
\begin{threeparttable}
\begin{tabular}{@{}l|rrr|rrr@{}}
\toprule
           & \multicolumn{3}{c|}{validation}                                                & \multicolumn{3}{c}{test}                                                      \\ 
           & mAP & AP50 & AP25 & mAP & AP50 & AP25 \\ \midrule
MTML~\cite{lahoud20193d}       & 20.3                    & 40.2                     & 55.4                     & 28.2                    & 54.9                     & 73.1                     \\
3D-MPA~\cite{engelmann20203d}    & 35.3                    & 59.1                     & 72.4                     & 35.5                    & 61.1                     & 73.7                     \\
PointGroup~\cite{jiang2020pointgroup} & 34.8                    & 56.9                     & 71.3                     & 40.7                    & 63.6            & \textbf{77.8}            \\
OccuSeg~\cite{han2020occuseg}    & \textbf{44.2}           & \textbf{60.7}            & 71.9                     & \textbf{44.3}           & 63.4                     & 73.9                     \\
\midrule
Learnable margin\tnote{\textdagger}~\cite{neven2019instance}   & 28.1                    & 50.1                     & 70.1                     & -    & -     & -     \\
Proposed\tnote{\textdagger}  & 33.0                    & 57.1                     & \textbf{73.8}            & 39.6                    & \textbf{64.5}                     & 77.6                     \\ \bottomrule
\end{tabular}
\begin{tablenotes}
    \item[\textdagger] Implemeneted with MinkowskiNet
\end{tablenotes}
\vspace{-7pt}
\caption{\textbf{Results on ScanNet.} We list the results on both validation and hidden test sets of ScanNet. 
Note that due to the unique submission policy of ScanNet, we are unable to provide the results of learnable margin on the test set. 
On validation set, we improve mAP by $4.9\%$.}
\label{tbl:scannet}
\end{threeparttable}
\end{table}

\subsection{Ablation study and analysis}
\label{sec:ablation}
We conduct the ablation study on all categories of PartNet~\cite{mo2019partnet}, but we only list detailed values for the four largest categories in Table~\ref{tbl:abl-study}.
\begin{table*}[tb]
\centering
\small\addtolength{\tabcolsep}{-2.5pt}
\resizebox{\linewidth}{!}{%
\centering
\begin{tabular}{@{}cc|ccccc|cc|cccc|cccccccc@{}}
\toprule
 Ablation & Model & \rotatebox[origin=lB]{90}{Center} &
  \rotatebox[origin=lB]{90}{ExtDim} & \rotatebox[origin=lB]{90}{Prob} &
 \rotatebox[origin=lB]{90}{Aniso} & \rotatebox[origin=lB]{90}{Hetero} & \rotatebox[origin=lB]{90}{AllAvg} & $\Delta$ & \rotatebox[origin=lB]{90}{Large} & $\Delta$  & \rotatebox[origin=lB]{90}{Others} & $\Delta$ & \rotatebox[origin=lB]{90}{Chair} & $\Delta$ & \rotatebox[origin=lB]{90}{Lamp} & $\Delta$ & \rotatebox[origin=lB]{90}{Stora} & $\Delta$ & \rotatebox[origin=lB]{90}{Table} & $\Delta$ \\ \midrule
 Loss &  & & & \checkmark & \checkmark & \checkmark & 54.3 & -0.7 & 16.2 & -7.3 & \textbf{43.1} & 3.1 & 19.0 & -8.4 & 8.8 & -9.6 & 29.7 & 6.2 & 7.1 & -17.4 \\ \midrule
\multirow{2}{*}{Deterministic} & Reference & \checkmark &  &  &  &  & 55.0 & 0.0 & 23.4 & 0.0 & 40.0 & 0.0 & 27.4 & 0.0 & 18.5 & 0.0 & 23.5 & 0.0 & 24.4 & 0.0 \\
 &  & \checkmark & \checkmark &  &  &  & \textbf{56.2} & 1.2 & 27.2 & 3.7 & 41.2 & 1.2 & 32.8 & 5.4 & \textbf{20.2} & 1.7 & 29.8 & 6.4 & \textbf{25.8} & 1.4 \\ \midrule
 \multirow{4}{*}{Probabilistic} &  & \checkmark  & & \checkmark &  &  & 54.7 & -0.3 & 26.4 & 3.0 & 39.9 & -0.1 & 33.7 & 6.3 & 17.8 & -0.7 & 30.0 & 6.5 & 24.3 & -0.2 \\
 &  & \checkmark &  & \checkmark &  & \checkmark & 53.1 & -1.9 & \textbf{27.4} & 3.9 & 38.1 & -1.8 & 33.6 & 6.2 & 20.0 & 1.5 & 30.9 & 7.5 & 25.0 & 0.6 \\
 &  & \checkmark &  & \checkmark & \checkmark &  & 55.6 & 0.6 & 27.3 & 3.9 & 41.0 & 1.1 & \textbf{33.9} & 6.5 & 19.5 & 1.0 & \textbf{31.1} & 7.6 & 24.8 & 0.3 \\ 
 & Full & \checkmark &  & \checkmark & \checkmark & \checkmark & \textbf{57.5} & 2.5 & \textbf{28.7} & 5.2 & \textbf{43.0} & 3.0 & \textbf{34.7} & 7.3 & \textbf{20.3} & 1.8 & \textbf{34.2} & 10.7 & \textbf{25.5} & 1.1 \\ \bottomrule
\end{tabular}
}
\vspace{-10pt}
\caption{\textbf{Ablation study}. \textbf{Center}, \textbf{ExtDim}, \textbf{Prob} refer to our proposed \em{center-aware} loss for the clustering step, the 6D deterministic embedding, and our proposed probabilistic embedding. \textbf{Aniso} and \textbf{Hetero} refer to the choice of Gaussian: anisotropic and heteroscedastic. \textbf{AllAvg} means taking all levels of granularity and categories into consideration. \textbf{Large} means fine-grained level of four largest categories. \textbf{Others} means fine-grained level of all the other categories. Here we also list the results on four largest categories of fine-grained level. The top two results are marked bold.}
\label{tbl:abl-study}
\end{table*}

\paragraph{Effect of probabilistic embedding.}
\begin{figure}[tb]
    \centering
    \input{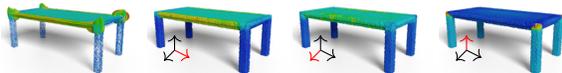}
    \vspace{-10pt}
    \caption{\textbf{Learned Uncertainties}. Top left: uncertainties are represented as ellipsoids, where directional scaling shows the value of uncertainties along 3 axes. The other 3 subfigures: uncertainties along 3 axes. We represent large values with red colors and smaller values with blue colors.}
    \label{fig:sigma}
\end{figure}
\begin{figure}
    \centering
    \includegraphics[width=1\linewidth]{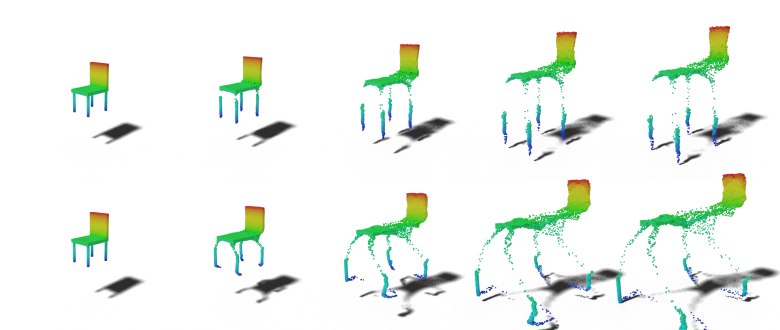}
    \vspace{-20pt}
    \caption{\textbf{Comparison of embeddings}. \textbf{Top row}: Deterministic embedding. \textbf{Bottom row}: Probabilistic embedding. \textbf{Left to right}: we show a gradual shape transformation between the original point cloud and the embedded point cloud.}
    \label{fig:offset}
\end{figure}
We compare four different versions of probabilistic embedding. The Gaussian distribution used in the model can either be isotropic or anisotropic, homoscedastic or heteroscedastic. Thus we have isotropic homoscedastic, anisotropic homoscedastic, isotropic heteroscedastic, and anisotropic heteroscedastic.

The isotropic homoscedastic probabilistic embedding, learns to predict a single scalar representing the uncertainty of a point cloud. We do not see improvements over its determinisitc counterpart, but there is a large gap between them in large categories which have much more part instances and classes than others.

Similar cases happen in anisotropic homoscedastic and isotropic homoscedastic embedding. The former learns a 3D uncertainty vector for a single point cloud, while the latter learns point-dependent uncertainty scalars. They all show significant improvements over determinisitc embedding on fine-grained categories.

Finally, our full model uses anisotropic heteroscedastic probabilistic embedding, which outputs not only point-dependent but also axis-dependent uncertainties. See Figure~\ref{fig:sigma} for an illustration of learned uncertainties. The points at boundary regions have significantly larger uncertainties compared to others. In summary, the full model achieves the best results among all variations.

\paragraph{Effect of spatial embedding.}
Since our full model outputs a 3D center vector and 3D uncertainty vector, in a way, we can regard it as a 6D embedding method (with a totally different similarity kernel). One may wonder: how does it compare with the performance of 6D deterministic embedding? The results in Table~\ref{tbl:abl-study} show, increasing the dimension of deterministic embedding from 3 to 6 shows some improvement, but less than using probabilistic embedding. Thus the performance of our method, cannot be achieved by simply increasing the dimension of deterministic embedding, which also shows the superiority of the probabilistic embedding. 
We illustrate the differences between deterministic and probabilistic embedding in 3D in Figure~\ref{fig:offset}. We can observe, that probabilistic embedding introduces much stronger deformations of the geometry.

\paragraph{Effect of center-aware loss.}
We examine the effect of the center-aware loss in the clustering step. We use the same setup as in our full model except changing the center-aware loss to MSE loss~\cite{neven2019instance}. In Table~\ref{tbl:abl-study}, we can see that our proposed loss function is especially stable on large fine-grained datasets (5.2\% vs -7.3\%).
\section{Conclusion}
We build on embedding-based instance segmentation to present a framework of probabilistic embedding and a new loss function for the clustering step. We evaluate our framework on a large scale point cloud dataset, PartNet, and achieve state-of-the-art performance. Moreover, the qualitative results show the new framework is robust to point clouds with many instances. Additionally, it is able to estimate uncertainties while increasing the accuracy of instance segmentation. In future work, we hope that the probabilistic embedding can be further applied to other kinds of data representation, \eg, 2D images, 3D volumes, and meshes.

\section*{Acknowledgements}
This work was supported by the KAUST Office of Sponsored Research (OSR) under Award No. OSR-CRG2017-3426.
\newpage
{\small
\bibliographystyle{ieee_fullname}
\bibliography{bib}
}

\section{Network architecture}
Following the notation of PointNet++~\cite{qi2017pointnet++}, we give the architecture of the feature network:
\begin{align*}
    & SA(512, 0.2, [64, 64, 128]), \\
    & SA(128, 0.4, [128, 128, 256]),\\
    & SA([256, 512, 1024]),\\
    & FP(256, 256), \\
    & FP(256, 128), \\
    & FP(128, 128), \\
\end{align*}
where $SA$ and $FP$ are \emph{set abstraction} and \emph{feature propagation} module in PointNet++~\cite{qi2017pointnet++}.
The output head network is:
\begin{align*}
    & FullyConnected(128, 256), \\
    & BatchNorm(256),\\
    & ReLU(), \\
    & FullyConnected(256, 128),\\
    & BatchNorm(128), \\
    & ReLU(), \\
    & FullyConnected(128, 128), \\
    & BatchNorm(128), \\
    & ReLU(), \\
    & FullyConnected(128, C). \\
\end{align*}

\section{Implementation}
We implemented our method using PyTorch~\cite{paszke2019pytorch} and the geometric deep learning library PyTorch Geometric~\cite{Fey/Lenssen/2019}. The final objective function is
\begin{equation}
    \mathcal{L} = \mathcal{L}_{Ins} + \mathcal{L}_{Score} + 0.001 \cdot \mathcal{L}_{Reg}
\end{equation}

\section{Results with different IoU thresholds}
We report detailed results of IoU using thresholds of 25\% and 75\% in Table~\ref{tbl:iou25} and Table~\ref{tbl:iou75}. The metric is mean Average Precision (mAP).
\begin{table*}[]
\centering
\small\addtolength{\tabcolsep}{-4.7pt}
\resizebox{\textwidth}{!}{%
\begin{tabular}{@{}l|c|l|cccccccccccccccccccccccc@{}}
\toprule
 &     & Avg & \rotatebox[origin=lB]{90}{Bag}  & \rotatebox[origin=lB]{90}{Bed}  & \rotatebox[origin=lB]{90}{Bottle} & \rotatebox[origin=lB]{90}{Bowl} & \rotatebox[origin=lB]{90}{Chair} & \rotatebox[origin=lB]{90}{Clock} & \rotatebox[origin=lB]{90}{Dish} & \rotatebox[origin=lB]{90}{Disp} & \rotatebox[origin=lB]{90}{Door} & \rotatebox[origin=lB]{90}{Ear}  & \rotatebox[origin=lB]{90}{Faucet} & \rotatebox[origin=lB]{90}{Hat}  & \rotatebox[origin=lB]{90}{Key}  & \rotatebox[origin=lB]{90}{Knife} & \rotatebox[origin=lB]{90}{Lamp} & \rotatebox[origin=lB]{90}{Laptop}   & \rotatebox[origin=lB]{90}{Micro} & \rotatebox[origin=lB]{90}{Mug}  & \rotatebox[origin=lB]{90}{Fridge} & \rotatebox[origin=lB]{90}{Scis} & \rotatebox[origin=lB]{90}{Stora} & \rotatebox[origin=lB]{90}{Table} & \rotatebox[origin=lB]{90}{Trash} & \rotatebox[origin=lB]{90}{Vase} \\ \midrule

  \multirow{4}{*}{\rotatebox[origin=c]{90}{PartNet}} & 1 & 70.2 & \textbf{89.4} & \textbf{82.3} & 65.2 & 63.1 & 78.1 & 48.0 & 79.1 & 97.1 & 64.9 & 64.6 & 77.3 & 73.9 & 58.9 & 59.2 & 42.5 & \textbf{100.0} & 50.0 & 92.9 & 50.0 & 96.3 & 57.7 & \textbf{59.3} & 82.7 & \textbf{52.6} \\
 & 2 & 46.7 & - & 44.5 & - & - & 43.0 & - & 71.3 & - & \textbf{49.3} & - & - & - & - & - & 32.2 & - & 51.2 & - & 45.2 & - & 46.7 & 36.5 & - & - \\
 & 3 & 45.6 & - & 29.0 & 52.6 & - & 35.3 & 39.6 & 59.9 & 89.3 & 27.1 & 56.9 & 55.0 & - & - & 49.0 & 22.6 & - & 56.9 & - & 35.6 & - & 36.3 & 28.6 & 44.8 & \textbf{57.0} \\ \cmidrule{2-27}
 & Avg & 62.8 & \textbf{89.4} & 51.9 & 58.9 & 63.1 & 52.1 & 43.8 & 70.1 & 93.2 & 47.1 & 60.8 & 66.2 & 73.9 & 58.9 & 54.1 & 32.4 & \textbf{100.0} & 52.7 & 92.9 & 43.6 & 96.3 & 46.9 & \textbf{41.5} & 63.8 & \textbf{54.8} \\ \midrule
\multirow{4}{*}{\rotatebox[origin=c]{90}{Ours}} & 1 & \textbf{72.7} & 82.8 & 79.6 & \textbf{65.6} & \textbf{72.0} & \textbf{82.8} & \textbf{49.1} & \textbf{83.8} & \textbf{98.3} & \textbf{75.5} & \textbf{74.3} & \textbf{83.2} & \textbf{79.5} & \textbf{59.9} & \textbf{78.8} & \textbf{45.2} & \textbf{100.0} & \textbf{50.5} & \textbf{95.4} & \textbf{51.6} & \textbf{96.9} & \textbf{60.9} & 44.6 & \textbf{82.9} & 51.1 \\
 & 2 & \textbf{51.4} & - & \textbf{55.4} & - & - & \textbf{47.1} & - & \textbf{78.0} & - & 48.1 & - & - & - & - & - & \textbf{39.3} & - & \textbf{54.4} & - & \textbf{48.8} & - & \textbf{53.7} & \textbf{37.7} & - & - \\
 & 3 & \textbf{51.6} & - & \textbf{44.4} & \textbf{57.2} & - & \textbf{43.2} & \textbf{45.7} & \textbf{64.8} & \textbf{90.7} & \textbf{34.6} & \textbf{59.3} & \textbf{67.2} & - & - & \textbf{53.0} & \textbf{26.0} & - & \textbf{60.0} & - & \textbf{51.5} & - & \textbf{44.4} & \textbf{31.7} & \textbf{50.0} & 53.9 \\ \cmidrule{2-27}
 & Avg & \textbf{66.5} & 82.8 & \textbf{59.8} & \textbf{61.4} & \textbf{72.0} & \textbf{57.7} & \textbf{47.4} & \textbf{75.6} & \textbf{94.5} & \textbf{52.7} & \textbf{66.8} & \textbf{75.2} & \textbf{79.5} & \textbf{59.9} & \textbf{65.9} & \textbf{36.8} & \textbf{100.0} & \textbf{55.0} & \textbf{95.4} & \textbf{50.6} & \textbf{96.9} & \textbf{53.0} & 38.0 & \textbf{66.5} & 52.5 \\ \bottomrule
\end{tabular}
}
\caption{\textbf{Instance segmentation results on PartNet}. The metric is mAP (\%) with IoU threshold 0.25.}
\label{tbl:iou25}
\end{table*}
\begin{table*}[]
\centering
\small\addtolength{\tabcolsep}{-4.7pt}
\resizebox{\textwidth}{!}{%
\begin{tabular}{@{}l|c|l|cccccccccccccccccccccccc@{}}
\toprule
 &     & Avg & \rotatebox[origin=lB]{90}{Bag}  & \rotatebox[origin=lB]{90}{Bed}  & \rotatebox[origin=lB]{90}{Bottle} & \rotatebox[origin=lB]{90}{Bowl} & \rotatebox[origin=lB]{90}{Chair} & \rotatebox[origin=lB]{90}{Clock} & \rotatebox[origin=lB]{90}{Dish} & \rotatebox[origin=lB]{90}{Disp} & \rotatebox[origin=lB]{90}{Door} & \rotatebox[origin=lB]{90}{Ear}  & \rotatebox[origin=lB]{90}{Faucet} & \rotatebox[origin=lB]{90}{Hat}  & \rotatebox[origin=lB]{90}{Key}  & \rotatebox[origin=lB]{90}{Knife} & \rotatebox[origin=lB]{90}{Lamp} & \rotatebox[origin=lB]{90}{Laptop}   & \rotatebox[origin=lB]{90}{Micro} & \rotatebox[origin=lB]{90}{Mug}  & \rotatebox[origin=lB]{90}{Fridge} & \rotatebox[origin=lB]{90}{Scis} & \rotatebox[origin=lB]{90}{Stora} & \rotatebox[origin=lB]{90}{Table} & \rotatebox[origin=lB]{90}{Trash} & \rotatebox[origin=lB]{90}{Vase} \\ \midrule
  \multirow{4}{*}{\rotatebox[origin=c]{90}{PartNet}} & 1 & 47.4 & 39.7 & \textbf{14.6} & 60.6 & 41.4 & 58.3 & \textbf{28.8} & 58.3 & 84.7 & 35.6 & 49.1 & 48.2 & 66.3 & \textbf{10.7} & 48.7 & \textbf{29.6} & \textbf{98.0} & 47.8 & 76.1 & 50.0 & 35.1 & 29.9 & \textbf{43.2} & \textbf{42.2} & 40.5 \\
 & 2 & 22.0 & - & 4.2 & - & - & 21.4 & - & 37.2 & - & \textbf{22.4} & - & - & - & - & - & 19.6 & - & 32.1 & - & 16.7 & - & 22.8 & 22.0 & - & - \\
 & 3 & 23.5 & - & 3.9 & 37.9 & - & 16.6 & \textbf{17.6} & 29.8 & 63.2 & 8.1 & 27.6 & 25.8 & - & - & 31.0 & 13.6 & - & 23.9 & - & 12.1 & - & 18.2 & 16.4 & \textbf{19.7} & 34.5 \\ \cmidrule{2-27}
 & Avg & 38.9 & 39.7 & 7.6 & 49.2 & 41.4 & 32.1 & \textbf{23.2} & 41.7 & 73.9 & 22.0 & 38.4 & 37.0 & 66.3 & \textbf{10.7} & 39.8 & 20.9 & \textbf{98.0} & 34.6 & 76.1 & 26.3 & 35.1 & 23.6 & \textbf{27.2} & \textbf{31.0} & 37.5 \\ \midrule
 \multirow{4}{*}{\rotatebox[origin=c]{90}{Ours}} & 1 & \textbf{50.0} & \textbf{40.3} & 13.3 & \textbf{60.2} & \textbf{60.2} & \textbf{59.3} & 28.2 & \textbf{61.9} & \textbf{90.6} & \textbf{39.1} & \textbf{59.6} & \textbf{54.2} & \textbf{69.3} & 7.4 & \textbf{65.7} & 28.5 & \textbf{98.0} & \textbf{47.9} & \textbf{77.1} & \textbf{50.5} & \textbf{42.8} & \textbf{30.1} & 34.8 & 40.7 & \textbf{41.1} \\
 & 2 & \textbf{23.8} & - & \textbf{7.1} & - & - & 22.8 & - & \textbf{37.4} & - & 21.3 & - & - & - & - & - & \textbf{22.0} & - & \textbf{35.5} & - & \textbf{20.6} & - & \textbf{26.1} & \textbf{21.4} & - & - \\
 & 3 & \textbf{25.7} & - & \textbf{7.3} & \textbf{38.8} & - & \textbf{20.5} & 17.2 & \textbf{30.0} & \textbf{66.8} & \textbf{10.8} & \textbf{28.2} & \textbf{33.2} & - & - & \textbf{31.5} & \textbf{14.1} & - & \textbf{25.6} & - & \textbf{17.1} & - & \textbf{21.0} & \textbf{17.4} & 19.4 & \textbf{38.0} \\ \cmidrule{2-27}
 & Avg & \textbf{41.7} & \textbf{40.3} & \textbf{9.2} & \textbf{49.5} & \textbf{60.2} & \textbf{34.2} & 22.7 & \textbf{43.1} & \textbf{78.7} & \textbf{23.7} & \textbf{43.9} & \textbf{43.7} & \textbf{69.3} & 7.4 & \textbf{48.6} & \textbf{21.5} & \textbf{98.0} & \textbf{36.4} & \textbf{77.1} & \textbf{29.4} & \textbf{42.8} & \textbf{25.7} & 24.5 & 30.0 & \textbf{39.6} \\ \bottomrule
\end{tabular}
}
\caption{\textbf{Instance segmentation results on PartNet}. The metric is mAP (\%) with IoU threshold 0.75.}
\label{tbl:iou75}
\end{table*}

\section{Qualitive Results}
\begin{figure*}[!htbp]
    \centering
    \includegraphics[width=\textwidth]{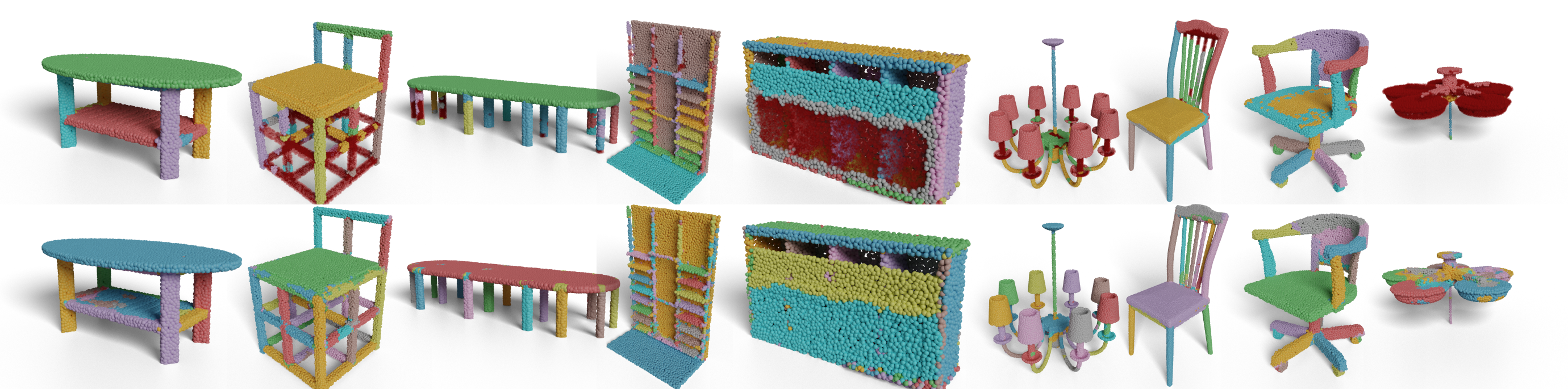}
    \caption{
        \textbf{\emph{All} detected instances:} unclassified points are shown in transparent red.
        \textbf{Top row}: PartNet. \textbf{Bottom row}: Ours.
        PartNet can group instance points together but fails to give the correct class labels in some cases 
        (\eg, in the first and the third subfigures from left to right, points of table legs are grouped together (top row) but they are not true positives. 
        Besides, in the sixth subfigure from left to right, PartNet fails to distinguish different instances of lamp covers. 
        Our method performs clearly better in these cases.}
    \label{fig:results-supp}
\end{figure*}

We present more qualitive results in Figure~\ref{fig:results-supp} which shows the instance-awareness of our method. We also demonstrate the 3D models in the attached video.

\section{Differences to learnable margin}
\cite{neven2019instance} proposed to use a learnable margin for image instance segmentation, which is similar in formulation to our proposed probabilistic embedding. Although we differ in several aspects:

\begin{enumerate}
    \item The intuition behind learnable margin comes from the hinge loss: to give different hinge margin to objects of different sizes. However, our intuition comes from modeling neural network outputs as random variables to estimate uncertainty.
    \item The parameters have a different meaning in our method compared to \cite{neven2019instance}. In learnable margin, $\sigma$ is an instance-specific bandwidth (or margin) per cluster. In our work $\sigma$ are uncertainties per point. 
    \item The bandwidth $\sigma$ is influenced by the size of instances (large instances have large $\sigma$). In contrast, our uncertainty $\sigma$ encodes per-point uncertainty close to the boundary of instances 
    (see Fig.~\ref{fig:sigma}).
    \item \cite{neven2019instance} add a loss term to enforce the bandwidths from the same instance to be close. By contrast, we don't have this kind of restriction. Also, uncertaintes from the same instance can be different as along as they have similar spatial embeddings.
\end{enumerate}
\end{document}